\pdfoutput=1
\documentclass{article}
\usepackage{ijcai20}

\usepackage{times}
\usepackage{soul}
\usepackage{url}
\usepackage[hidelinks]{hyperref}
\usepackage[utf8]{inputenc}
\usepackage[small]{caption}
\usepackage{graphicx}
\usepackage{amsmath}
\usepackage{amsthm}
\usepackage{booktabs}
\usepackage{algorithm}
\usepackage{algorithmic}
\usepackage{amssymb}
\usepackage{xcolor}
\usepackage{multirow}
\usepackage{booktabs}
\usepackage{pifont}

\usepackage{tikz}
\usepackage{subfigure}
\usepackage{pgfplots}

\pgfplotsset{compat=1.14}

\usetikzlibrary{plotmarks}
\usetikzlibrary{arrows}
\usetikzlibrary{decorations}
\usetikzlibrary{decorations.pathreplacing}
\usetikzlibrary{decorations.markings}
\usetikzlibrary{backgrounds}
\usetikzlibrary{fit}
\usetikzlibrary{positioning}
\usetikzlibrary{calc}
\usetikzlibrary{patterns}
\usetikzlibrary{shadows}
\usetikzlibrary[shapes.multipart]
\usetikzlibrary{pgfplots.groupplots}

\newdimen\base
\newdimen\baseh
\newdimen\basew
\baseh=\base
\basew=\base

\newdimen\legendmargin 
\newdimen\legendwidth 
\newdimen\legendsep 

\definecolor{ugreen}{rgb}{0,0.5,0}
\definecolor{ublue}{rgb}{0.152,0.250,0.545}
\definecolor{lyyblue}{RGB}{31,120,180}
\definecolor{lyygreen}{RGB}{51,160,44}
\definecolor{lyyred}{RGB}{227,26,28}

\tikzset{
    ncbar angle/.initial=90,
    ncbar/.style={
        to path=(\tikztostart)
        -- ($(\tikztostart)!#1!\pgfkeysvalueof{/tikz/ncbar angle}:(\tikztotarget)$)
        -- ($(\tikztotarget)!($(\tikztostart)!#1!\pgfkeysvalueof{/tikz/ncbar angle}:(\tikztotarget)$)!\pgfkeysvalueof{/tikz/ncbar angle}:(\tikztostart)$)
        -- (\tikztotarget)
    },
    ncbar/.default=0.5cm,
}

\newcommand{\citet}[1]{\citeauthor{#1} \shortcite{#1}}
\newcommand{\specialcell}[4][c]{%
  #3\begin{tabular}[#1]{@{}#2@{}}#4\end{tabular}}
\newcommand{\softmax}{$\mathrm{SoftMax}$}
\newcommand{\relu}{$\mathrm{ReLU}$}

\title{Towards Fully 8-bit Integer Inference for the Transformer Model}

\author{
Ye Lin$^1$\thanks{Authors contributed equally.}\and
Yanyang Li$^{1*}$\and
Tengbo Liu$^1$\and
Tong Xiao$^{1,2}$\thanks{Corresponding author.}\and
Tongran Liu$^3$\And
Jingbo Zhu$^{1,2}$\\
\affiliations
$^1$Natural Language Processing Lab., Northeastern University, Shenyang, China\\
$^2$NiuTrans Research, Shenyang, China\\
$^3$CAS Key Laboratory of Behavioral Science, Institute of Psychology, CAS, Beijing, China
\emails
\{linye2015, blamedrlee\}@outlook.com,
tengboliu@stumail.neu.edu.cn,\\
\{xiaotong, zhujingbo\}@mail.neu.edu.cn,
liutr@psych.ac.cn
}

\begin{document}

\maketitle

\begin{abstract}
    8-bit integer inference, as a promising direction in reducing both the latency and storage of deep neural networks, has made great progress recently. On the other hand, previous systems still rely on 32-bit floating point for certain functions in complex models (e.g., Softmax in Transformer), and make heavy use of quantization and de-quantization. In this work, we show that after a principled modification on the Transformer architecture, dubbed \emph{Integer Transformer}, an (almost) fully 8-bit integer inference algorithm \emph{Scale Propagation} could be derived. De-quantization is adopted when necessary, which makes the network more efficient. Our experiments on WMT16 En$\leftrightarrow$Ro, WMT14 En$\leftrightarrow$De and En$\rightarrow$Fr translation tasks as well as the WikiText-103 language modelling task show that the fully 8-bit Transformer system achieves comparable performance with the floating point baseline but requires nearly 4$\times$ less memory footprint.
\end{abstract}

\section{Introduction}
\label{sec:introduction}

    In recent years, the self-attention-based Transformer model \cite{DBLP:conf/nips/VaswaniSPUJGKP17} has shown promising improvements in a wide variety of tasks, e.g., machine translation \cite{li2020aaai} and language modelling \cite{DBLP:conf/iclr/BaevskiA19}. The superior performance of these systems is mostly achieved by using very large neural networks, which are accompanied by the great demands on computation, storage and energy \cite{DBLP:conf/acl/StrubellGM19}. As a side effect, deploying such models on small devices is challenging as they have limited storage space and computation power. For example, practical systems often run on CPUs where the 32-bit floating point computation capability is much lower than that of GPUs.

    One appealing solution to these issues is to reduce the numerical precision used in the model at hand \cite{DBLP:conf/nips/HubaraCSEB16,DBLP:conf/iclr/MicikeviciusNAD18}, where both the parameters and the activations are represented with fewer bits. For instance, employing 8-bit integer (INT8) potentially consumes 4$\times$ less storage space but is up to 6$\times$ faster \cite{DBLP:conf/naacl/QuinnB18}. Beyond this, INT8 is 10$\times$ more energy efficient \cite{DBLP:journals/corr/abs-1811-01721} and saves much less chip area than the commonly used 32-bit floating point (FP32) in hardware design \cite{DBLP:journals/pieee/SzeCYE17}. Also, the low-precision approach is orthogonal to other existing compression and acceleration methods, e.g., efficient network design \cite{DBLP:conf/ijcai/XiaoLZ0L19}.

    \begin{figure}[t!]
        \subfigure[Ideal INT8 Inference]
        {
            \begin{tikzpicture}
                \tikzstyle{opnode} = [rectangle,draw,rounded corners=2pt,minimum width=0.9\base,minimum height=0.5\base,font=\scriptsize,align=center,inner sep=3pt]
                \tikzstyle{labelnode} = [font=\scriptsize,inner sep=1pt]

                \baseh=0.8\base
                \basew=\base

                \node [labelnode] (input) at (0,0) {$\left\{x,s\right\}$};
                \node [opnode,anchor=west,fill=blue!30!white] (op1) at ([xshift=0.83\base]input.east) {$\mathrm{OP}1$};
                \node [labelnode,anchor=west] (output1) at ([xshift=\base]op1.east) {$\left\{x',s'\right\}$};
                \node [opnode,anchor=west,fill=blue!30!white] (op2) at ([xshift=1.1\base]output1.east) {$\mathrm{OP}2$};
                \node [labelnode,anchor=west] (output2) at ([xshift=0.65\base]op2.east) {$\left\{x'',s''\right\}$};

                \draw [-latex'] (input) to (op1);
                \draw [-latex'] (op1) to (output1);
                \draw [-latex'] (output1) to (op2);
                \draw [-latex'] (op2) to (output2);
            \end{tikzpicture}
        }\\[0pt]
        \centering
        \subfigure[Practical INT8 Inference]
        {
            \begin{tikzpicture}
                \tikzstyle{opnode} = [rectangle,draw,rounded corners=2pt,minimum width=0.9\base,minimum height=0.5\base,font=\scriptsize,align=center,inner sep=3pt]
                \tikzstyle{labelnode} = [font=\scriptsize,inner sep=1pt]

                \baseh=0.3\base
                \basew=0.25\base

                \node [labelnode] (input) at (0,0) {$r$};
                \node [opnode,anchor=west,fill=ugreen!40!white] (quant1) at ([xshift=\basew]input.east) {$Q()$};
                \node [opnode,anchor=west,fill=blue!30!white] (op1) at ([xshift=\basew]quant1.east) {$\mathrm{OP}1$};
                \node [opnode,anchor=west,fill=red!30!white] (dequant1) at ([xshift=\basew]op1.east) {$D()$};
                \node [labelnode,anchor=west] (output1) at ([xshift=\basew]dequant1.east) {$r'$};
                \node [opnode,anchor=west,fill=ugreen!40!white] (quant2) at ([xshift=\basew]output1.east) {$Q()$};
                \node [opnode,anchor=west,fill=blue!30!white] (op2) at ([xshift=\basew]quant2.east) {$\mathrm{OP}2$};
                \node [opnode,anchor=west,fill=red!30!white] (dequant2) at ([xshift=\basew]op2.east) {$D()$};
                \node [labelnode,anchor=west] (output2) at ([xshift=\basew]dequant2.east) {$r''$};

                \draw [-latex'] (input) to (quant1);
                \draw [-latex'] (quant1) to (op1);
                \draw [-latex'] (op1) to (dequant1);
                \draw [-latex'] (dequant1) to (output1);
                \draw [-latex'] (output1) to (quant2);
                \draw [-latex'] (quant2) to (op2);
                \draw [-latex'] (op2) to (dequant2);
                \draw [-latex'] (dequant2.east) to ([xshift=\basew]dequant2.east);
            \end{tikzpicture}
        }
        \caption{Ideal vs. Practical INT8 inference ($\mathrm{OP}$: operation).}
        \label{fig:example}
    \end{figure}
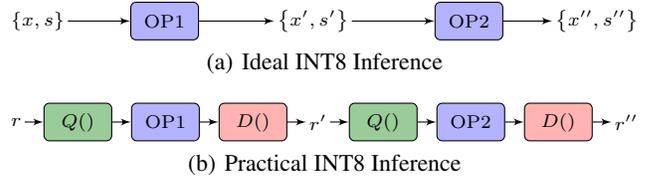

    In general, we need two additional components to adapt FP32 algorithms to INT8 algorithms: quantization and de-quantization \cite{DBLP:conf/asplos/GongSZLLJMFS18}. Quantization can be seen as a function that transforms a rational tensor $r$ into an integer tensor $x$ with the scale $s$ \cite{DBLP:journals/corr/abs-2001-00926}:
    \begin{equation}
        Q(r,s)=\left\lfloor s \cdot r \right\rceil
        \label{eqn:quant}
    \end{equation}
    where $\left\lfloor \cdot \right\rceil$ represents rounding to the nearest integer. As a reverse process, de-quantization approximates the rational tensor $r$ with its quantized form $x$:
    \begin{equation}
        D(x,s)=x/s
        \label{eqn:dequant}
    \end{equation}

    Ideally, the INT8-based inference process is as follow: the rational input (FP32) tensor $r$ is first quantized to an INT8 tensor $x$ with the scale $s$. Then all succeeding operations are performed on INT8 tensors and corresponding scales simultaneously. De-quantization is employed at the end of the process or there appears an overflow\footnote{For intermediate tensors produced by these operations, we perform de-quantization and quantization with $s=\frac{2^p-1}{\max(|r|)}$ immediately if the overflow happens. The bit-precision $p$ is 7 for INT8.}.

    This method is efficient because quantization and de-quantization functions are used only when necessary. Unfortunately, previous INT8-based models are much more expensive, as every operation in it is sandwiched between a pair of quantization and de-quantization (see Fig. \ref{fig:example}). The heavy use of quantization and de-quantization blocks the efficient flow of INT8 throughout the network, and somehow prevents fully 8-bit integer models. The problem lies in two facts:
    \begin{itemize}
        \item Scale Incompatibility: INT8 tensors with different scales are incomparable because we cannot use the same FP32-to-INT8 mapping to process them in a single operation. For example, let $x_1$ and $x_2$ be  INT8 tensors that are quantized from FP32 tensors $r_1$ and $r_2$ with difference scales $s_1$ and $s_2$. Adding $x_1$ and $x_2$ is obviously problematic because $x_1+x_2$ is not the INT8 form of $r_1 + r_2$, i.e., $r_1+r_2 \neq (x_1+x_2)/s_1 \neq (x_1+x_2)/s_2$.
        \item INT8 Incompatibility: some functions in complex networks are not INT8 friendly and we have to resort to FP32 computation in this case. The most representative examples are the exponential function in the attention mechanism and the square root function in layer normalization \cite{DBLP:conf/nips/VaswaniSPUJGKP17}.
    \end{itemize}

    In this work, we take a further step towards fully INT8-based  transformer models. We choose Transformer for study because it is one of the most popular models in natural language processing. We present \emph{Scale Propagation}, which bounds INT8 tensors with associated scales, and propagates them throughout the network during inference. It addresses the scale incompatibility issue by matching the input scales if necessary, allowing each operation to manipulate the INT8 tensor and its scale simultaneously. Moreover, we propose \emph{Integer Transformer} in responding to the INT8 incompatibility issue. To make full use of INT8 in Transformer, we replace the exponential function in the standard attention by the polynomial function, and replace the square root function in the layer normalization with the absolute value function. Our extensive experiments on several machine translation and language modelling tasks show that integer Transformer achieves competitive INT8 performance with approximately 4$\times$ less storage and 3.47$\times$ speed-up on average.

    \section{Background: Transformer}

    We start with the description of Transformer. Transformer \cite{DBLP:conf/nips/VaswaniSPUJGKP17} is mainly composed of a stack of layers. Each layer consists of a self-attention and a feed-forward network. The self-attention takes three tensors, $Q$, $K$ and $V$, as inputs and produces a tensor with the same size as the output. It is formulated as:
    \begin{equation}
        \mathrm{Attention}(Q,K,V)=\mathrm{SoftMax}(\frac{QK^T}{\sqrt{d_\mathrm{m}}})V
        \label{eqn:attention}
    \end{equation}
    where $d_m$ is the dimension of the hidden representation. \softmax{} is a function that casts its input to a distribution:
    \begin{equation}
        \mathrm{SoftMax}(x_i)=\frac{e^{x_i}}{\sum_j e^{x_j}}
        \label{eqn:softmax}
    \end{equation}

    The feed-forward network is built on top of two linear projections with the \relu{} activation in between:
    \begin{eqnarray}
        \mathrm{FFN}(x)&= &\mathrm{ReLU}(xW_1+b_1)W_2+b_2 \label{eqn:ffn} \\
        \mathrm{ReLU}(x)&= &\max(0,x) \label{eqn:relu}
    \end{eqnarray}

    These modules are coupled with the residual connection \cite{DBLP:conf/cvpr/HeZRS16}, i.e., $y=f(x)+x$ where $f$ is either the self-attention or the feed-forward network. The Layer Normalization is after the residual connection:
    \begin{equation}
        \mathrm{LN}(x)=g \odot (\frac{x- \mu }{\sqrt{\sigma^2+ \varepsilon }})+b
        \label{eqn:layernorm}
    \end{equation}
    where $\mu$ and  $\sigma$ are the mean and variance of $x$ along the hidden dimension, and $\varepsilon$ is a fixed small number to prevent dividing 0. $g$ and $b$ are two learnable parameters. For more details, we refer the reader to \cite{DBLP:conf/nips/VaswaniSPUJGKP17}.

\section{Scale Propagation}

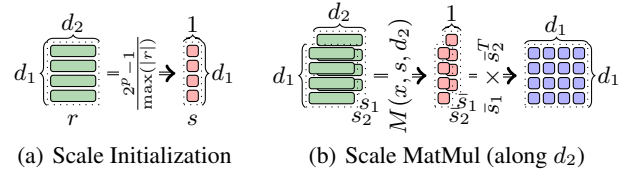
\begin{figure}[t!]
        \centering
        \hspace*{\fill}
        \subfigure[Scale Initialization]
        {
            \begin{tikzpicture}
                \tikzstyle{vecnode} = [rectangle,draw,rounded corners=1pt,minimum height=0.15\base,minimum width=0.6\base,inner sep=0pt]
                \tikzstyle{labelnode} = [font=\footnotesize,inner sep=0pt]

                \baseh=0.3\base
                \basew=1.2\base

                \coordinate (start) at (0,0);
                \node [vecnode,fill=ugreen!30!white,anchor=south west] (r1) at (start) {};
                \node [vecnode,fill=ugreen!30!white,anchor=south west] (r2) at ([yshift=1pt]r1.north west) {};
                \node [vecnode,fill=ugreen!30!white,anchor=south west] (r3) at ([yshift=1pt]r2.north west) {};
                \node [vecnode,fill=ugreen!30!white,anchor=south west] (r4) at ([yshift=1pt]r3.north west) {};
                \node [draw,dotted,inner sep=1pt,fit=(r1) (r4),label={[]-90:\footnotesize$r$}] (r) {};
                \draw [decorate,decoration={brace}] (r.north west) to node [auto] {\footnotesize$d_2$} (r.north east);
                \draw [decorate,decoration={brace,mirror}] (r.north west) to node [auto,swap] {\footnotesize$d_1$} (r.south west);

                \node [vecnode,minimum width=0.15\base,fill=red!30!white,anchor=west] (s1) at ([xshift=\basew]r1.east) {};
                \node [vecnode,minimum width=0.15\base,fill=red!30!white,anchor=west] (s2) at ([xshift=\basew]r2.east) {};
                \node [vecnode,minimum width=0.15\base,fill=red!30!white,anchor=west] (s3) at ([xshift=\basew]r3.east) {};
                \node [vecnode,minimum width=0.15\base,fill=red!30!white,anchor=west] (s4) at ([xshift=\basew]r4.east) {};
                \node [draw,dotted,inner sep=1pt,fit=(s1) (s4),label={[]-90:\footnotesize$s$}] (s) {};
                \draw [decorate,decoration={brace}] (s.north west) to node [auto] {\footnotesize$1$} (s.north east);
                \draw [decorate,decoration={brace}] (s.north east) to node [auto] {\footnotesize$d_1$} (s.south east);

                \draw [->,double] ([xshift=3pt]r.east) to node [midway,rotate=90,fill=white,inner sep=1pt,font=\footnotesize,yshift=1pt] {$\frac{2^p-1}{\max(|r|)}$} ([xshift=-3pt]s.west);
            \end{tikzpicture}
            \label{fig:init-scale}
        }
        \hfill
        \subfigure[Scale MatMul (along $d_2$)]
        {
            \begin{tikzpicture}
                \tikzstyle{vecnode} = [rectangle,draw,rounded corners=1pt,minimum height=0.15\base,minimum width=0.6\base,inner sep=0pt]
                \tikzstyle{labelnode} = [font=\footnotesize,inner sep=0pt]

                \baseh=1\base
                \basew=1.1\base

                \coordinate (start) at (0,0);
                \node [vecnode,fill=ugreen!30!white,anchor=south west] (s11) at (start) {};
                \node [vecnode,fill=ugreen!30!white,anchor=south west] (s12) at ([yshift=1pt]s11.north west) {};
                \node [vecnode,fill=ugreen!30!white,anchor=south west] (s13) at ([yshift=1pt]s12.north west) {};
                \node [vecnode,fill=ugreen!30!white,anchor=south west] (s14) at ([yshift=1pt]s13.north west) {};
                \node [draw,dotted,inner sep=1pt,fit=(s11) (s14),label={[label distance=-2pt,font=\footnotesize]-70:$s_1$}] (s1) {};
                \draw [decorate,decoration={brace}] (s1.north west) to node [auto] {\footnotesize$d_2$} (s1.north east);

                \coordinate (start) at ([shift={(-3pt,-5pt)}]s11.south west);
                \node [vecnode,fill=ugreen!30!white,anchor=south west] (s21) at (start) {};
                \node [vecnode,fill=ugreen!30!white,anchor=south west] (s22) at ([yshift=1pt]s21.north west) {};
                \node [vecnode,fill=ugreen!30!white,anchor=south west] (s23) at ([yshift=1pt]s22.north west) {};
                \node [vecnode,fill=ugreen!30!white,anchor=south west] (s24) at ([yshift=1pt]s23.north west) {};
                \node [draw,dotted,inner sep=1pt,fit=(s21) (s24),label={[label distance=-2pt,font=\footnotesize]-70:$s_2$}] (s2) {};
                \draw [decorate,decoration={brace,mirror}] (s2.north west) to node [auto,swap] {\footnotesize$d_1$} (s2.south west);

                \node [vecnode,minimum width=0.15\base,fill=red!30!white,anchor=west] (s31) at ([xshift=\basew]s11.east) {};
                \node [vecnode,minimum width=0.15\base,fill=red!30!white,anchor=west] (s32) at ([xshift=\basew]s12.east) {};
                \node [vecnode,minimum width=0.15\base,fill=red!30!white,anchor=west] (s33) at ([xshift=\basew]s13.east) {};
                \node [vecnode,minimum width=0.15\base,fill=red!30!white,anchor=west] (s34) at ([xshift=\basew]s14.east) {};
                \node [draw,dotted,inner sep=1pt,fit=(s31) (s34)] (s3) {};
                \draw [decorate,decoration={brace}] (s3.north west) to node [auto] {\footnotesize$1$} (s3.north east);
                \node [labelnode,anchor=north west] () at ([xshift=-2pt]s3.south east) {$\bar{s}_1$};

                \node [vecnode,minimum width=0.15\base,fill=red!30!white,anchor=west] (s41) at ([xshift=\basew]s21.east) {};
                \node [vecnode,minimum width=0.15\base,fill=red!30!white,anchor=west] (s42) at ([xshift=\basew]s22.east) {};
                \node [vecnode,minimum width=0.15\base,fill=red!30!white,anchor=west] (s43) at ([xshift=\basew]s23.east) {};
                \node [vecnode,minimum width=0.15\base,fill=red!30!white,anchor=west] (s44) at ([xshift=\basew]s24.east) {};
                \node [draw,dotted,inner sep=1pt,fit=(s41) (s44)] (s4) {};
                \node [labelnode,anchor=north west] () at ([xshift=-2pt]s4.south east) {$\bar{s}_2$};

                \coordinate (s510) at ([xshift=\baseh]s41.east);
                \foreach \cur [count=\prev from 0] in {1,2,...,4}
                    \node [vecnode,minimum width=0.15\base,fill=blue!30!white,anchor=west] (s51\cur) at ([xshift=1pt]s51\prev.east) {};
                \coordinate (s520) at ([xshift=\baseh]s42.east);
                \foreach \cur [count=\prev from 0] in {1,2,...,4}
                    \node [vecnode,minimum width=0.15\base,fill=blue!30!white,anchor=west] (s52\cur) at ([xshift=1pt]s52\prev.east) {};
                \coordinate (s530) at ([xshift=\baseh]s43.east);
                \foreach \cur [count=\prev from 0] in {1,2,...,4}
                    \node [vecnode,minimum width=0.15\base,fill=blue!30!white,anchor=west] (s53\cur) at ([xshift=1pt]s53\prev.east) {};
                \coordinate (s540) at ([xshift=\baseh]s44.east);
                \foreach \cur [count=\prev from 0] in {1,2,...,4}
                    \node [vecnode,minimum width=0.15\base,fill=blue!30!white,anchor=west] (s54\cur) at ([xshift=1pt]s54\prev.east) {};
                \node [draw,dotted,inner sep=1pt,fit=(s511) (s544)] (s5) {};
                \draw [decorate,decoration={brace}] (s5.north west) to node [auto] {\footnotesize$d_1$} (s5.north east);
                \draw [decorate,decoration={brace}] (s5.north east) to node [auto] {\footnotesize$d_1$} (s5.south east);

                \draw [->,double] ([xshift=6pt]s2.east) to node [midway,rotate=90,font=\footnotesize,fill=white,inner sep=0pt,yshift=0pt] {$M(x,s,d_2)$} ([xshift=-3pt]s4.west);
                \draw [->,double] ([xshift=6pt]s4.east) to node [midway,rotate=90,font=\footnotesize,fill=white,inner sep=0pt,yshift=0pt,pos=0.45] {$\bar{s}_1\times\bar{s}^T_2$} ([xshift=-3pt]s5.west);
            \end{tikzpicture}
            \label{fig:matmul-scale}
        }
        \hspace*{\fill}
        \caption{Examples of initializing scale and multiplying scale in MatMul.}
        \label{fig:scale}
    \end{figure}

\subsection{Bounding Tensors \& Scales}
\label{sec:bounding}

    As discussed in Section \ref{sec:introduction}, the necessity of de-quantization comes from the fact that input INT8 tensors might not be produced by the same mapping that converts FP32 to INT8, e.g., not multiplied by the same scale in our case, and therefore forces us to compute the correct result by rolling back to the FP32 mode.

    Inspired by this fact, the ideal INT8-based inference should propagate not only the tensor but also the mapping. If multiple INT8 tensors are inputted, unifying their mappings is necessary so that we can perform the succeeding operation on INT8 tensors directly. In our case, the mapping is defined as the scale that indicates to what extent the current INT8 tensor deviates from its FP32 counterpart. This scale can be obtained through $s=\frac{2^p-1}{\max(|r|)}$ initially, where $r$ is the network input. In practice, $\max(|r|)$ is performed along the hidden dimension, producing a scale with the same shape as $r$ except the dimension that the maximization operates is 1. Fig. \ref{fig:scale}\ref{sub@fig:init-scale} shows how to initialize the scale from an FP32 tensor. Although we introduce extra operations to manipulate FP32 scales, it is cheap to maintain them and the cost is negligible.

    \begin{table}[t!]
        \centering
        \begin{tabular}{r|l}
            \toprule
            \multicolumn{1}{c|}{FP32 $\mathrm{OP}$} & \multicolumn{1}{c}{INT8 Equivalent} \\
            \midrule
            $\left[r_1,r_2\right]$ & $\{\left[x_1, x_2\right],\left[s_1, s_2\right]\}$ \\
            $r^T_1$ & $\{x_1^T,s_1^T\}$ \\
            \midrule
            $r_1 \cdot r_2$ & $\{x_1 \cdot x_2,s_1 \cdot s_2\}$ \\
            $r_1+r_2$ & $\{x_i,\bar{s}\}=M(x_i,s_i),i\in\{1,2\}$\\
            & $\{x_1 + x_2,\bar{s}\},\bar{s} \in \mathbb{R}^{m \times n}$ \\
            $\textrm{MatMul}(r_1, r^T_2)$ & $\{x_i,\bar{s}_i\}=M(x_i,s_i,d_2),i\in\{1,2\}$\\
            & $\{x_1 \times x^T_2,\bar{s}_1 \times \bar{s}^T_2\},\bar{s}_i \in \mathbb{R}^{m \times 1}$ \\
            \midrule
            $r^n_1$ & $\{x_1^n,s_1^n\}$ \\
            $|r_1|$ & $\{|x_1|,s_1\}$ \\
            $\mathrm{ReLU}(r_1)$ & $\{\mathrm{ReLU}(x_1),s_1\}$ \\
            \bottomrule
        \end{tabular}
        \caption{FP32 operations in INT8. $r_i=D(x_i,s_i)$, $i \in \{1,2\}$, $r_i \in \mathbb{R}^{m \times n}$, $x_i \in \mathbb{Z}^{m \times n}$, $s_i \in \mathbb{R}^{m \times n}$. $\left[\right]$ denotes the concatenation. $\cdot$ denotes the element-wise multiplication.}
        \label{tab:op}
    \end{table}

\subsection{Manipulating Tensors \& Scales}

    Extending the common FP32 operations to INT8 tensors and associated scales is non-trivial. Two questions naturally arise: 1) how can we calibrate mappings? 2) how to operate both the INT8 tensors and scales for a given FP32 operation?

    For the first question, we note that the mapping here is a scale that gets multiplied in the quantization. Thus having an identical mapping across tensors is as to find a unique multiplier for every input tensor in our case. For each input tensor $x_i$ with the scale $s_i$, we do \emph{Scale Matching}:
    \begin{equation}
        M(x_i,s_i)=\{x_i / \left\lceil s_i/\bar{s} \right\rceil,\bar{s}\}
        \label{eqn:scale-match}
    \end{equation}
    where $\bar{s}=\min(s_1,\cdots,s_n)$. Choosing the minimum of scales $\bar{s}$ as the unique multiplier guarantees that the result of Eq. \ref{eqn:scale-match} does not overflow.

    Having the scale matching, it is handy to induce the INT8 form for any FP32 tensor operation, as shown in Table \ref{tab:op}. For tensor shape transformations, such as concatenation and transpose, the same transformation is applied to the INT8 tensor and its scale simultaneously, since they do not change the values. For element-wise multiplication, we multiply tensors and scales independently, as the quantization is just another element-wise multiplication. For addition, we first match the input scales via Eq. \ref{eqn:scale-match}, then add tensors as usual.

    Handling matrix multiplication (MatMul) is more sophisticated. MatMul is an element-wise multiplication with an addition along the last dimension. We therefore first match the input scales along that dimension, then perform MatMul to the tensors and scales independently. To match the input scale along a specific dimension, we employ the same idea of scale matching by treating it as matching scales of multiple sub-tensors splitted from that dimension. It is denoted as $M(x,s,d)$, where $x$ is the INT8 tensor, $s$ is its scale and $d$ is the dimension that we would like to match scales. Fig. \ref{fig:scale}\ref{sub@fig:matmul-scale} shows an example of how MatMul works on scales, which matches the scales on the dimension $d_2$ and multiplies them.

    \begin{algorithm}[t!]
        \caption{\textsc{Scale Propagation Protocol}}
        \label{alg:scale-prop}
        \textbf{Input}: Operation $\mathrm{OP}$; INT8 Tensors $x_{1 \ldots n}$; Scales $s_{1 \ldots n}$ \\
        \textbf{Output}: INT8 Tensor $x$; Scale $s$
        \begin{algorithmic}[1]
            \STATE $\{x,s\}=\mathrm{OP}(\{x_{1 \ldots n},s_{1 \ldots n}\})$ \COMMENT{Store $x$ in INT32}
            \IF{$x>2^p-1$}
            \STATE $\{x,s\}=R(x,s)$ \COMMENT{Re-scaling}
            \ENDIF
            \STATE Convert (INT32) $x$ to INT8
            \RETURN $x,s$
        \end{algorithmic}
    \end{algorithm}

    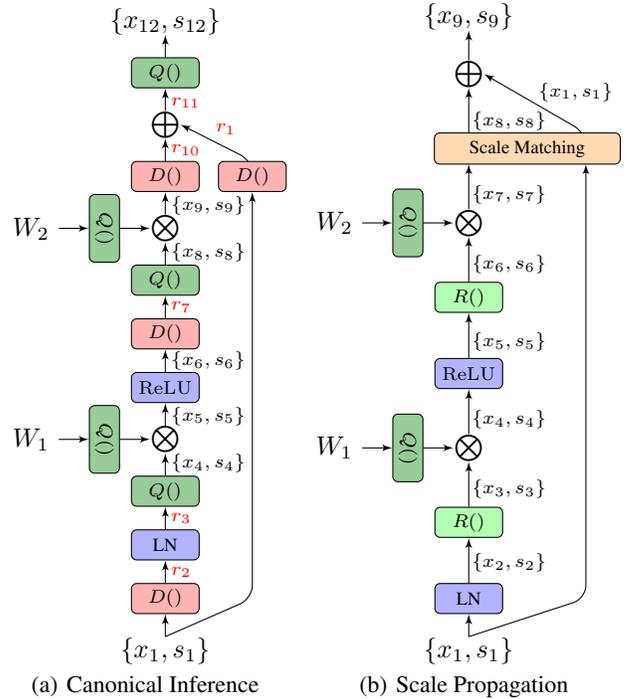
\begin{figure}[t!]
       \centering
       \hspace*{\fill}
       \subfigure[Canonical Inference]
       {
           \begin{tikzpicture}
               \tikzstyle{opnode} = [rectangle,draw,rounded corners=2pt,minimum width=0.9\base,minimum height=0.4\base,font=\scriptsize,align=center,inner sep=0pt]
               \tikzstyle{labelnode} = [font=\scriptsize,inner sep=0pt]

               \baseh=0.3\base
               \basew=0.4\base

               \begin{scope}
                   \node [font=\normalsize,inner sep=0pt] (input) at (0,0) {$\{x_1,s_1\}$};
                   \node [opnode,anchor=south,fill=red!30!white] (de1) at ([yshift=\baseh]input.north) {$D()$};
                   \node [opnode,anchor=south,fill=blue!30!white] (ln) at ([yshift=\baseh]de1.north) {LN};
                   \node [opnode,anchor=south,fill=ugreen!40!white] (quant1) at ([yshift=\baseh]ln.north) {$Q()$};
                   \node [anchor=south,inner sep=0pt] (matmul1) at ([yshift=\baseh]quant1.north) {$\bigotimes$};
                   \node [opnode,anchor=south,fill=blue!30!white] (relu) at ([yshift=\baseh]matmul1.north) {\relu{}};
                   \node [opnode,anchor=south,fill=red!30!white] (de2) at ([yshift=\baseh]relu.north) {$D()$};
                   \node [opnode,anchor=south,fill=ugreen!40!white] (quant2) at ([yshift=\baseh]de2.north) {$Q()$};
                   \node [anchor=south,inner sep=0pt] (matmul2) at ([yshift=\baseh]quant2.north) {$\bigotimes$};
                   \node [opnode,anchor=south,fill=red!30!white] (de3) at ([yshift=\baseh]matmul2.north) {$D()$};
                   \node [anchor=south,inner sep=0pt] (add) at ([yshift=\baseh]de3.north) {$\bigoplus$};
                   \node [opnode,anchor=south,fill=ugreen!40!white] (quant5) at ([yshift=\baseh]add.north) {$Q()$};
                   \node [font=\normalsize,inner sep=0pt,anchor=south] (output) at ([yshift=0.3\base]quant5.north) {$\{x_{12},s_{12}\}$};

                   \draw [-latex'] (input) to (de1);
                   \draw [-latex'] (de1) to (ln);
                   \draw [-latex'] (ln) to (quant1);
                   \draw [-latex'] (quant1) to (matmul1);
                   \draw [-latex'] (matmul1) to (relu);
                   \draw [-latex'] (relu) to (de2);
                   \draw [-latex'] (de2) to (quant2);
                   \draw [-latex'] (quant2) to (matmul2);
                   \draw [-latex'] (matmul2) to (de3);
                   \draw [-latex'] (de3) to (add);
                   \draw [-latex'] (add) to (quant5);
                   \draw [-latex'] (quant5) to (output);

                   \node [opnode,rotate=-90,anchor=north,fill=ugreen!40!white] (quant3) at ([xshift=-\basew]matmul1.west) {$Q()$};
                   \node [font=\normalsize,anchor=east] (w1) at ([xshift=-\basew]quant3.south) {$W_1$};
                   \node [opnode,rotate=-90,anchor=north,fill=ugreen!40!white] (quant4) at ([xshift=-\basew]matmul2.west) {$Q()$};
                   \node [font=\normalsize,anchor=east] (w2) at ([xshift=-\basew]quant4.south) {$W_2$};

                   \draw [-latex'] (quant3) to (matmul1);
                   \draw [-latex'] (quant4) to (matmul2);
                   \draw [-latex'] (w1) to (quant3);
                   \draw [-latex'] (w2) to (quant4);

                   \coordinate (corner) at ([xshift=0.7\base]de1.east);
                   \node [opnode,fill=red!30!white] (de4) at ([xshift=0.7\base]de3.east) {$D()$};
                   \draw [-latex',rounded corners=5pt] (input.north) to (corner) to (de4.south);\begin{pgfonlayer}{background}
                       \draw [-latex',rounded corners=5pt] (de4.south) to (de4.north) to (add.east);
                   \end{pgfonlayer}

                   \node [labelnode,anchor=south west,text=red] () at ([shift={(2pt,2pt)}]de1.north) {$r_2$};
                   \node [labelnode,anchor=south west,text=red] () at ([shift={(2pt,2pt)}]ln.north) {$r_3$};
                   \node [labelnode,anchor=south west] () at ([shift={(2pt,1pt)}]quant1.north) {$\{x_4,s_4\}$};
                   \node [labelnode,anchor=south west] () at ([shift={(2pt,0pt)}]matmul1.north) {$\{x_5,s_5\}$};
                   \node [labelnode,anchor=south west] () at ([shift={(2pt,0.5pt)}]relu.north) {$\{x_6,s_6\}$};
                   \node [labelnode,anchor=south west,text=red] () at ([shift={(2pt,2pt)}]de2.north) {$r_7$};
                   \node [labelnode,anchor=south west] () at ([shift={(2pt,1pt)}]quant2.north) {$\{x_8,s_8\}$};
                   \node [labelnode,anchor=south west] () at ([shift={(2pt,0pt)}]matmul2.north) {$\{x_9,s_9\}$};
                   \node [labelnode,anchor=south west,text=red] () at ([shift={(2pt,2pt)}]de3.north) {$r_{10}$};
                   \node [labelnode,anchor=south west,text=red] () at ([shift={(-14pt,10pt)}]de4.north) {$r_{1}$};
                   \node [labelnode,anchor=south west,text=red] () at ([shift={(2pt,1pt)}]add.north) {$r_{11}$};
               \end{scope}
           \end{tikzpicture}
           \label{fig:canonical}
       }
       \hfill
       \subfigure[Scale Propagation]
       {
           \begin{tikzpicture}
               \tikzstyle{opnode} = [rectangle,draw,rounded corners=2pt,font=\scriptsize,align=center,inner sep=0pt,minimum width=0.9\base,minimum height=0.4\base]
               \tikzstyle{labelnode} = [font=\scriptsize,inner sep=0pt]

               \baseh=0.6\base
               \basew=0.4\base

               \begin{scope}
                   \node [font=\normalsize,inner sep=0pt] (input) at (0,0) {$\{x_1,s_1\}$};
                   \node [opnode,anchor=south,fill=blue!30!white] (ln) at ([yshift=0.3\base]input.north) {LN};
                   \node [opnode,anchor=south,fill=green!30!white] (quant1) at ([yshift=\baseh]ln.north) {$R()$};
                   \node [anchor=south,inner sep=0pt] (matmul1) at ([yshift=\baseh]quant1.north) {$\bigotimes$};
                   \node [opnode,anchor=south,fill=blue!30!white] (relu) at ([yshift=\baseh]matmul1.north) {\relu{}};
                   \node [opnode,anchor=south,fill=green!30!white] (quant2) at ([yshift=\baseh]relu.north) {$R()$};
                   \node [anchor=south,inner sep=0pt] (matmul2) at ([yshift=\baseh]quant2.north) {$\bigotimes$};
                   \node [opnode,anchor=south,opacity=0] (match1) at ([yshift=\baseh]matmul2.north) {};
                   \node [anchor=south,inner sep=0pt] (add) at ([yshift=\baseh]match1.north) {$\bigoplus$};
                   \node [font=\normalsize,inner sep=0pt,anchor=south] (output) at ([yshift=0.4\base]add.north) {$\{x_9,s_9\}$};

                   \draw [-latex'] (input) to (ln);
                   \draw [-latex'] (ln) to (quant1);
                   \draw [-latex'] (quant1) to (matmul1);
                   \draw [-latex'] (matmul1) to (relu);
                   \draw [-latex'] (relu) to (quant2);
                   \draw [-latex'] (quant2) to (matmul2);
                   \draw [-latex'] (matmul2) to (match1);
                   \draw [-latex'] (match1) to (add);
                   \draw [-latex'] (add) to (output);

                   \node [opnode,rotate=-90,anchor=north,fill=ugreen!40!white] (quant3) at ([xshift=-\basew]matmul1.west) {$Q()$};
                   \node [font=\normalsize,anchor=east] (w1) at ([xshift=-\basew]quant3.south) {$W_1$};
                   \node [opnode,rotate=-90,anchor=north,fill=ugreen!40!white] (quant4) at ([xshift=-\basew]matmul2.west) {$Q()$};
                   \node [font=\normalsize,anchor=east] (w2) at ([xshift=-\basew]quant4.south) {$W_2$};

                   \draw [-latex'] (quant3) to (matmul1);
                   \draw [-latex'] (quant4) to (matmul2);
                   \draw [-latex'] (w1) to (quant3);
                   \draw [-latex'] (w2) to (quant4);

                   \coordinate (corner) at ([xshift=1.1\base]ln.east);
                   \node [opnode,opacity=0] (match2) at ([xshift=1.1\base]match1.east) {};
                   \draw [-latex',rounded corners=5pt] (input.north) to (corner) to (match2.south);
                   \begin{pgfonlayer}{background}
                       \draw [-latex',rounded corners=5pt] (match2.south) to (match2.north) to (add.east);
                   \end{pgfonlayer}

                   \node [opnode,fill=orange!30!white,fit=(match1) (match2)] (match) {};
                   \node [font=\scriptsize] () at (match.center) {Scale Matching};

                   \node [labelnode,anchor=south west] () at ([shift={(1pt,4pt)}]ln.north) {$\{x_2,s_2\}$};
                   \node [labelnode,anchor=south west] () at ([shift={(1pt,3pt)}]quant1.north) {$\{x_3,s_3\}$};
                   \node [labelnode,anchor=south west] () at ([shift={(1pt,2pt)}]matmul1.north) {$\{x_4,s_4\}$};
                   \node [labelnode,anchor=south west] () at ([shift={(1pt,2.5pt)}]relu.north) {$\{x_5,s_5\}$};
                   \node [labelnode,anchor=south west] () at ([shift={(1pt,3pt)}]quant2.north) {$\{x_6,s_6\}$};
                   \node [labelnode,anchor=south west] () at ([shift={(1pt,2pt)}]matmul2.north) {$\{x_7,s_7\}$};
                   \node [labelnode,anchor=south west] () at ([shift={(1pt,1pt)}]match1.north) {$\{x_8,s_8\}$};
                   \node [labelnode,anchor=south west] () at ([shift={(-18pt,12pt)}]match2.north) {$\{x_1,s_1\}$};
               \end{scope}
           \end{tikzpicture}
           \label{fig:scale-prop}
       }
       \hspace*{\fill}
       \caption{The comparison of INT8 inferences in the FFN layer.}
       \label{fig:inference}
   \end{figure}

    For element-wise non-linear functions, we assume that they satisfy the \emph{distribution law}, i.e., $\mathrm{OP}(r)=\mathrm{OP}(x/s)=\mathrm{OP}(x)/\mathrm{OP}(s)$. Then, we have:
    \begin{equation}
        \mathrm{OP}(\{x,s\})=\{\mathrm{OP}(x),\mathrm{OP}(s)\}
        \label{eqn:quant-element}
    \end{equation}
    where $x$ is the INT8 tensor and $s$ is its scale. This assumption holds for the polynomial function $r^n$ where $n$ is a fixed integer, since $r^n=(x/s)^n=x^n/s^n$. It also holds for the absolute value function, because $|r|=|x/s|=|x|/s$ as $s>0$. The same is for $\mathrm{ReLU}(\{x,s\})$ when entries of $x$ have the maximum value(s) exceeded 0, which is always true otherwise it will face the `dying ReLU' problem \cite{DBLP:conf/iccv/HeZRS15}.

    \begin{figure*}[t!]
        \centering
        \begin{tikzpicture}

            \tikzset{square left brace/.style={ncbar=0.1cm}}
            \tikzset{square right brace/.style={ncbar=-0.1cm}}

            \tikzstyle{wordnode} = [font=\normalsize,draw,fill=yellow!10!white,rectangle,rounded corners=3pt,drop shadow]
            \tikzstyle{probnode} = [rectangle,minimum width=0.3\base,font=\scriptsize,inner sep=0pt]

            \baseh=0.5\base
            \basew=\base

            \begin{scope}
                \node [wordnode] (v) at (0,0) {$V$};
                \node [wordnode,anchor=south] (k) at ([yshift=0.8\base]v.north) {$K$};
                \node [wordnode,anchor=south] (q) at ([yshift=0.8\base]k.north) {$Q$};

                \node [wordnode,anchor=west] (qk) at ([xshift=1.75\basew]k.east) {${\color{blue}x}=\frac{{\color{red}Q}{\color{ugreen}K^T}}{\sqrt{d_m}}$};
                \node [wordnode,anchor=west] (poly) at ([xshift=2.5\basew]qk.east) {${\color{blue}x}=\mathrm{Poly}({\color{blue}x})$};
                \node [wordnode,anchor=west] (sum) at ([xshift=2.5\basew]poly.east) {${\color{orange}x}=\sum_j{\color{blue}x_j}$};
                \node [wordnode,anchor=west] (sumv) at ([xshift=8.25\base]v.east) {${\color{red}y}={\color{blue}x}{\color{ugreen}V}$};
                \node [wordnode,anchor=west] (div) at ([xshift=3.25\base]sumv.east) {${\color{red}y}={\color{red}y}/{\color{orange}x}$};

                \begin{pgfonlayer}{background}
                    \path [draw=red!30,line width=5pt,rounded corners=5pt] (q.east) -- ([xshift=1.25\base]q.east) -- (qk.west);
                    \path [draw=ugreen!30,line width=5pt] (k.east) to (qk.west);
                    \path [draw=blue!30,line width=5pt] (qk.east) to (poly.west);
                    \path [draw=blue!30,line width=5pt] (poly.east) to (sum.west);
                    \path [draw=ugreen!30,line width=5pt] (v.east) to (sumv.west);
                    \path [draw=blue!30,line width=5pt,rounded corners=5pt] (poly.east) to ([xshift=0.75\base]poly.east) to ([xshift=0.25\base]sumv.north);
                    \path [draw=red!30,line width=5pt] (sumv.east) to (div.west);
                    \path [draw=orange!30,line width=5pt,rounded corners=5pt] (sum.east) to ([xshift=0.75\base]sum.east) to (div.north);
                    \path [draw=red!30,line width=5pt] (div.east) to ([xshift=1.5\base]div.east);

                    \draw [-latex',thick,rounded corners=5pt] (q.east) to ([xshift=1.25\base]q.east) to (qk.west);
                    \draw [-latex',thick] (k.east) to (qk.west);

                    \draw [-latex',thick] (qk.east) to (poly.west);

                    \draw [-latex',thick] (poly.east) to (sum.west);

                    \draw [-latex',thick] (v.east) to (sumv.west);
                    \draw [-latex',thick,rounded corners=5pt] (poly.east) to ([xshift=0.75\base]poly.east) to ([xshift=0.25\base]sumv.north);

                    \draw [-latex',thick] (sumv.east) to (div.west);
                    \draw [-latex',thick,rounded corners=5pt] (sum.east) to ([xshift=0.75\base]sum.east) to (div.north);

                    \draw [-latex',thick] (div.east) to ([xshift=1.5\base]div.east);
                \end{pgfonlayer}

                \coordinate (start) at ([shift={(1.65\base,-0.2\base)}]q.east);
                \node [rectangle,draw,rounded corners=1pt,minimum height=0.1\base,minimum width=0.8\base,fill=red!30!white,inner sep=0pt,anchor=south west] (p1) at (start) {};
                \node [draw,dotted,inner sep=1pt,fit=(p1)] (mat) {};
                \draw [decorate,decoration={brace}] (mat.north west) to node [auto] {\footnotesize$d_m$} (mat.north east);
                \draw [decorate,decoration={brace}] (mat.north east) to node [auto] {\footnotesize$1$} (mat.south east);

                \coordinate (start) at ([shift={(0.2\base,0.2\base)}]k.east);
                \node [rectangle,draw,rounded corners=1pt,minimum height=0.1\base,minimum width=0.8\base,fill=ugreen!30!white,inner sep=0pt,anchor=south west] (p1) at (start) {};
                \node [rectangle,draw,rounded corners=1pt,minimum height=0.1\base,minimum width=0.8\base,fill=ugreen!30!white,inner sep=0pt,anchor=south west] (p2) at ([yshift=1pt]p1.north west) {};
                \node [rectangle,draw,rounded corners=1pt,minimum height=0.1\base,minimum width=0.8\base,fill=ugreen!30!white,inner sep=0pt,anchor=south west] (p3) at ([yshift=1pt]p2.north west) {};
                \node [rectangle,draw,rounded corners=1pt,minimum height=0.1\base,minimum width=0.8\base,fill=ugreen!30!white,inner sep=0pt,anchor=south west] (p4) at ([yshift=1pt]p3.north west) {};
                \node [draw,dotted,inner sep=1pt,fit=(p1) (p4)] (mat) {};
                \draw [decorate,decoration={brace}] (mat.north west) to node [auto] {\footnotesize$d_m$} (mat.north east);
                \draw [decorate,decoration={brace}] (mat.north east) to node [auto] {\footnotesize$4$} (mat.south east);

                \coordinate (start) at ([shift={(0.3\base,0.5\base)}]qk.east);
                \node [probnode,fill=blue!30!white,minimum height=0.3\base,anchor=north west] (p1) at ([xshift=1pt]start) {};
                \node [probnode,fill=blue!30!white,minimum height=0.2\base,anchor=south west] (p2) at ([xshift=1pt]p1.north east) {};
                \node [probnode,fill=blue!30!white,minimum height=0.6\base,anchor=south west] (p3) at ([xshift=1pt]p2.south east) {};
                \node [probnode,fill=blue!30!white,minimum height=0.5\base,anchor=south west] (p4) at ([xshift=1pt]p3.south east) {};
                \coordinate (end) at ([xshift=6pt]p4.south east);
                \draw [-latex,thick] (start) to (end) node [right,align=center] {$0$};

                \coordinate (start) at ([shift={(0.3\base,0.5\base)}]poly.east);
                \node [probnode,fill=blue!30!white,minimum height=0.07\base,anchor=south west] (p1) at ([xshift=1pt]start) {};
                \node [probnode,fill=blue!30!white,minimum height=0.07\base,anchor=south west] (p2) at ([xshift=1pt]p1.south east) {};
                \node [probnode,fill=blue!30!white,minimum height=0.9\base,anchor=south west] (p3) at ([xshift=1pt]p2.south east) {};
                \node [probnode,fill=blue!30!white,minimum height=0.4\base,anchor=south west] (p4) at ([xshift=1pt]p3.south east) {};
                \coordinate (end) at ([xshift=6pt]p4.south east);
                \draw [-latex,thick] (start) to (end) node [right,align=center] {$0$};
                \draw [blue,rounded corners=3pt,thick] ([yshift=0.1\base]p1.north) to ([yshift=0.1\base]p2.north) to ([yshift=0.1\base]p3.north) to ([yshift=0.1\base]p4.north);

                \coordinate (start) at ([shift={(0.3\base,0.5\base)}]sum.east);
                \node [rectangle,draw,rounded corners=1pt,minimum size=0.2\base,fill=orange!30!white,inner sep=0pt,anchor=south west] (p1) at (start) {};
                \node [draw,dotted,inner sep=1pt,fit=(p1)] (mat) {};
                \draw [decorate,decoration={brace}] (mat.north west) to node [auto] {\footnotesize$1$} (mat.north east);
                \draw [decorate,decoration={brace}] (mat.north east) to node [auto] {\footnotesize$1$} (mat.south east);

                \coordinate (start) at ([shift={(4\base,0.2\base)}]v.east);
                \node [rectangle,draw,rounded corners=1pt,minimum height=0.1\base,minimum width=0.8\base,fill=ugreen!30!white,inner sep=0pt,anchor=south west] (p1) at (start) {};
                \node [rectangle,draw,rounded corners=1pt,minimum height=0.1\base,minimum width=0.8\base,fill=ugreen!30!white,inner sep=0pt,anchor=south west] (p2) at ([yshift=1pt]p1.north west) {};
                \node [rectangle,draw,rounded corners=1pt,minimum height=0.1\base,minimum width=0.8\base,fill=ugreen!30!white,inner sep=0pt,anchor=south west] (p3) at ([yshift=1pt]p2.north west) {};
                \node [rectangle,draw,rounded corners=1pt,minimum height=0.1\base,minimum width=0.8\base,fill=ugreen!30!white,inner sep=0pt,anchor=south west] (p4) at ([yshift=1pt]p3.north west) {};
                \node [draw,dotted,inner sep=1pt,fit=(p1) (p4)] (mat) {};
                \draw [decorate,decoration={brace}] (mat.north west) to node [auto] {\footnotesize$d_m$} (mat.north east);
                \draw [decorate,decoration={brace}] (mat.north east) to node [auto] {\footnotesize$4$} (mat.south east);

                \coordinate (start) at ([shift={(1.75\base,0.3\base)}]sumv.east);
                \node [rectangle,draw,rounded corners=1pt,minimum height=0.1\base,minimum width=0.8\base,fill=red!30!white,inner sep=0pt,anchor=south west] (p1) at (start) {};
                \node [draw,dotted,inner sep=1pt,fit=(p1)] (mat) {};
                \draw [decorate,decoration={brace}] (mat.north west) to node [auto] {\footnotesize$d_m$} (mat.north east);
                \draw [decorate,decoration={brace}] (mat.north east) to node [auto] {\footnotesize$1$} (mat.south east);

                \coordinate (start) at ([shift={(0.35\base,0.3\base)}]div.east);
                \node [rectangle,draw,rounded corners=1pt,minimum height=0.1\base,minimum width=0.8\base,fill=red!30!white,inner sep=0pt,anchor=south west] (p1) at (start) {};
                \node [draw,dotted,inner sep=1pt,fit=(p1)] (mat) {};
                \draw [decorate,decoration={brace}] (mat.north west) to node [auto] {\footnotesize$d_m$} (mat.north east);
                \draw [decorate,decoration={brace}] (mat.north east) to node [auto] {\footnotesize$1$} (mat.south east);
            \end{scope}
        \end{tikzpicture}
        \caption{A running example of Polynomial Attention, where $Q \in \mathbb{R}^{1 \times d_m}$, $K,V \in \mathbb{R}^{4 \times d_m}$. Different colors indicate different shapes.}
        \label{fig:attention}
    \end{figure*}
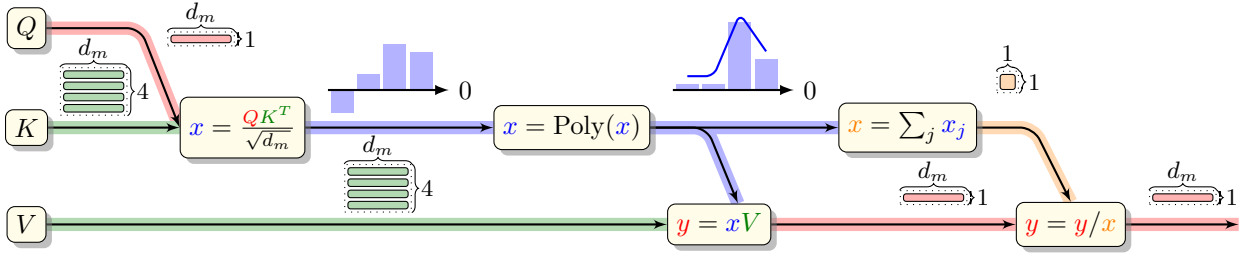

\subsection{The General Protocol}

    Note that addition and multiplication operations may produce results that are out of the INT8 range. These results are thereby stored in data types with more bits in practical implementations, e.g., INT32. We need to project the result back to INT8 before the succeeding operations. We call it \emph{Re-scaling}:
    \begin{equation}
        R(x,s)=\{x/\hat{s},s/\hat{s}\}
        \label{eqn:re-quant}
    \end{equation}

    \noindent where $\hat{s}=\left\lceil \frac{\max(|x|)}{2^p-1} \right\rceil$. The protocol of extending an FP32 operation to INT8 tensors and their scales is summarized in Alg. \ref{alg:scale-prop}: we directly apply the INT8 form of this operation to update $\{x,s\}$, and then use re-scaling to project $x$ back to INT8 if necessary. Once this protocol for INT8 operations is defined, the routine for the INT8 forward propagation is as straightforward as the FP32 one, except that FP32 operations are replaced by their INT8 equivalents. This gives us the \emph{Scale Propagation}. As shown in Fig. \ref{fig:inference}, scale propagation gets rid of de-quantization and only INT8 tensors are propagated in the whole forward propagation.

\section{Integer Transformer}

\subsection{Polynomial Attention}

    Applying scale propagation to the Transformer model is not immediately available. As discussed in the previous section, scale propagation assumes the element-wise functions satisfy the distribution law, which is not held for the exponential function $e^x$ in the \softmax{} of the attention functions, as $e^{x/s} \neq e^x/e^s$. Besides, the exponential function does not produce an integer output given an integer input.

    To enable scale propagation, we choose \relu{} as an alternative of the exponential function here, since it not only produces positive results as the exponential function but also is compatible with INT8. A bias term is added in advance to rule out entries that are below a learnt threshold.

    One downside of \relu{} is its linear nature. The exponential function has the property that larger input values become more significant after the transformation, as its gradient $e^x$ is larger than 1 in the positive number field. To achieve a similar effect, we introduce the polynomial function $x^n$, whose gradient $n \cdot x^{n-1}$ is exponential while it always produces integer outputs given integer inputs. Note that we only apply this polynomial function after \relu{}, since an even degree $n$ will mess up with the order of scores: a large negative number will be ranked in front instead of behind.

    Putting all these pieces together, we have:
    \begin{equation}
        \mathrm{Poly}(x)=\left[\mathrm{ReLU}(x+b)\right]^n+|\delta|
        \label{eqn:poly}
    \end{equation}
    where $b$ is the bias term, $n$ is the degree of the polynomial function and $\delta$ is another learnable parameter. $|\delta|$ ensures that the worst case of the attention, i.e., producing all $0$ results, is a simple average instead of nothing.

    Lastly, we multiply $\mathrm{Poly}(x)$ with $V$ and then divide the result by $\sum_j\mathrm{Poly}(x_j)$, otherwise the integer division will incur all 0 results because $\mathrm{Poly}(x_i)\le\sum_j\mathrm{Poly}(x_j)$:
    \begin{equation}
        \mathrm{PolyAttn}(Q,K,V)=\frac{\mathrm{Poly}(\frac{QK^T}{\sqrt{d_m}})V}{\sum_j\mathrm{Poly}(\frac{QK^T_j}{\sqrt{d_m}})}
        \label{eqn:poly-attn}
    \end{equation}
    This way sidesteps the previous issue as the multiplication results are usually not smaller than the sum. We call Eq. \ref{eqn:poly-attn} \emph{Polynomial Attention}. Fig. \ref{fig:attention} shows a running example of it.

\subsection{L1 Layer Normalization}

    Another component that hinders Transformer INT8 inference is the square root function for computing the standard deviation inside the layer normalization, which does not guarantee the integer outputs given the integer inputs. \citet{DBLP:conf/nips/HofferBGS18} proposes \emph{L1 Batch Normalization}, which approximates the standard deviation with its L1-norm equivalent:
    \begin{equation}
        \mathrm{L1LN}(x)=g\odot(\frac{x-\mu}{C\cdot\parallel x-\mu \parallel_1/n})+b
        \label{eqn:l1ln}
    \end{equation}
    where $g$ and $b$ are two parameters, $\mu$ is the mean of $x$ along the batch dimension, $C=\sqrt{\pi/2}$ and $n$ is the batch size. This way replaces the square root function in the L2-norm by the absolute value function in the L1-norm.

    We extend a similar idea of L1 batch normalization to our case, that we compute the mean $\mu$ along the hidden dimension instead of the dimension along the batch. We call this \emph{L1 Layer Normalization}. The replacement of layer normalization as well as the attention gives us the \emph{Integer Transformer} that supports fully INT8 inference.

\section{Experiments}

\subsection{Setup}

    We evaluate our methods on three machine translation (MT) tasks and a language modelling (LM) task, including the WMT16 English-Roman (En$\leftrightarrow$Ro), the WMT14 English-German (En$\leftrightarrow$De), the WMT14 English-French (En$\rightarrow$Fr) and the WikiText-103 LM tasks. For En$\leftrightarrow$Ro (610K pairs), we use \emph{newsdev-2016} and \emph{newstest-2016} as the validation and test sets respectively. For En$\leftrightarrow$De (4.5M pairs), \emph{newstest-2013} is the validation set and \emph{newstest-2014} is the test set. For En$\rightarrow$Fr (36M pairs), we validate the system on the combination of \emph{newstest-2012} and \emph{newstest-2013}, and test it on \emph{newstest-2014}. We tokenize every sentence using a script from Moses and segment every word into subword units using byte-pair encoding. The number of the BPE merge operations is set to 32K. We report case-sensitive tokenized BLEU scores. In addition, the results are the average of three identical runs with different random seeds for En$\leftrightarrow$Ro and En$\leftrightarrow$De. The WikiText-103 dataset contains a training set of 103 million words. Both the validation and test sets contain 0.2 million words. For the LM task, we report the perplexity.

    For the machine translation tasks, we experiment with the Transformer-base (base) setting \cite{DBLP:conf/acl/WangLXZLWC19}. We additionally run the Transformer-big (big) setting on En$\leftrightarrow$De and En$\rightarrow$Fr. Both settings consist of a 6-layer encoder and a 6-layer decoder. The embedding size is set to 512 for Transformer-base and 1,024 for Transformer-big. The number of heads is 8/16 for Transformer-base/big. The hidden size equals to 4$\times$ embedding size in both settings. For training, we use Adam optimizer with $\beta_1=0.9$ and $\beta_2=0.997$. We adopt the inverse square root learning rate schedule with 8K warmup steps and the learning rate $=0.001$/$0.0007$ for Transformer-base/big.

    For the language modelling task, we follow the lm-base and lm-big architectural choices and training details described in \cite{DBLP:conf/iclr/BaevskiA19}. The embedding size is 512 for lm-base and 1024 for lm-big. The hidden size equals to 4$\times$ embedding size. The number of heads is 8 for both lm-base and lm-big. The number of layers is set to 6/16 for lm-base/big. For the lm-base model, we train it with the same setting as in the machine translation tasks. As for the lm-big training, we use the Nesterov's accelerated gradient. We adopt the cosine learning rate schedule with 16K warmup steps and the maximum learning rate $1$. All experiments are run on 8 NVIDIA TITAN V GPUs.

    \begin{table}[t!]
        \centering
        \renewcommand\tabcolsep{4pt}
        \begin{tabular}{c|c|l|c|c|r|r}
            \hline
            \multicolumn{2}{c|}{\multirow{2}*{Entry}} & \multicolumn{1}{c|}{\multirow{2}*{System}} & \multicolumn{2}{c|}{BLEU} & \multicolumn{1}{c|}{\multirow{2}*{\small Storage}} & \multicolumn{1}{c}{\multirow{2}*{\specialcell{c}{\small}{Estimated\\Speed-up}}} \\
            \cline{4-5}
            \multicolumn{2}{c|}{} & & FP32 & INT8 & & \\
            \hline
            \multirow{10}*{\rotatebox{90}{base}} & \multirow{2}*{En$\rightarrow$Ro} & Baseline & 32.55 & - & 318M & $1 \times$ \\
            \cline{3-7}
            & & Ours & \textbf{32.60} & 32.54 & 80M & $3.53 \times$ \\
            \cline{2-7}
            & \multirow{2}*{Ro$\rightarrow$En} & Baseline & 32.85 & - & 306M & $1 \times$ \\
            \cline{3-7}
            & & Ours & \textbf{33.04} & 32.95 & 77M & $3.59 \times$ \\
            \cline{2-7}
            & \multirow{2}*{En$\rightarrow$De} & Baseline & 26.95 & - & 302M & $1 \times$ \\
            \cline{3-7}
            & & Ours & \textbf{27.08} & 26.91 & 76M & $3.24 \times$ \\
            \cline{2-7}
            & \multirow{2}*{De$\rightarrow$En} & Baseline & 32.19 & - & 302M & $1 \times$ \\
            \cline{3-7}
            & & Ours & \textbf{32.43} & 32.26 & 76M & $3.31 \times$ \\
            \cline{2-7}
            & \multirow{2}*{En$\rightarrow$Fr} & Baseline & \textbf{40.88} & - & 425M & $1 \times$ \\
            \cline{3-7}
            & & Ours & 40.64 & 40.00 & 107M & $3.03 \times$ \\
            \hline
            \multirow{6}*{\rotatebox{90}{big}} & \multirow{2}*{En$\rightarrow$De} & Baseline & 28.72 & - & 939M & $1 \times$ \\
            \cline{3-7}
            & & Ours & \textbf{28.93} & 28.71 & 236M & $3.60 \times$ \\
            \cline{2-7}
            & \multirow{2}*{De$\rightarrow$En} & Baseline & 33.07 & - & 939M & $1 \times$ \\
            \cline{3-7}
            & & Ours & \textbf{33.53} & 33.46 & 236M & $3.68 \times$ \\
            \cline{2-7}
            & \multirow{2}*{En$\rightarrow$Fr} & Baseline & 42.37 & - & 1243M & $1 \times$ \\
            \cline{3-7}
            & & Ours & \textbf{42.46} & 41.59 & 311M & $3.51 \times$ \\
            \hline
        \end{tabular}
        \caption{BLEU scores [\%], storage (megabytes) and speed-up.}
        \label{tab:mt}
    \end{table}

    \begin{table}[t!]
        \centering
        \renewcommand\tabcolsep{2.5pt}
        \begin{tabular}{c|l|c|c|c|c|r|r}
            \hline
            \multicolumn{2}{c|}{\multirow{2}*{Entry}} & \multicolumn{2}{c|}{valid} & \multicolumn{2}{c|}{test} & \multicolumn{1}{c|}{\multirow{2}*{\small Storage}} & \multicolumn{1}{c}{\multirow{2}*{\specialcell{c}{\small}{Estimated\\Speed-up}}} \\
            \cline{3-6}
            \multicolumn{2}{c|}{} & FP32 & INT8 & FP32 & INT8 & & \\
            \hline
            \multirow{2}*{\rotatebox{90}{base}} & Baseline & 29.61 & - & 31.18 & - & 596M & $1 \times$ \\
            \cline{2-8}
            & Ours & \textbf{29.49} & 30.28 & \textbf{30.79} & 31.61 & 150M & $3.43 \times$ \\
            \cline{1-8}
            \multirow{2}*{\rotatebox{90}{big}} & Baseline & 18.22 & - & 18.86 & - & 944M & $1 \times$ \\
            \cline{2-8}
            & Ours & \textbf{17.49} & 17.55 & \textbf{18.16} & 18.23 & 280M & $3.78 \times$ \\
            \hline
        \end{tabular}
        \caption{WikiText-103 PPL, storage (megabytes) and speed-up.}
        \label{tab:lm}
    \end{table}

\subsection{Results}

    Table \ref{tab:mt} summarizes the results on various translation tasks. Compared to the vanilla Transformer, integer Transformer obtains competitive or even superior FP32 performance by 0.1$\sim$0.4 BLEU points in either the base or big setup. When integer Transformer is decoded with INT8, it shows only about a decrease of 0.3 BLEU points on average except in En$\rightarrow$Fr, where it underperforms the baseline by more than 1 BLEU point. In Section \ref{sec:analysis}, we will show that it is mainly due to the last residual connection and layer normalization, which suffer from greater loss with lower bits representations. Experiments on the WikiText-103 language modelling task in Table \ref{tab:lm} show a similar trend as those in machine translation tasks, where integer Transformer beats the baseline with the same setup as in MT.

    Both Table \ref{tab:mt} and Table \ref{tab:lm} show that using INT8 indeed saves nearly 4$\times$ storage space. Since we need to store both the parameters and their scales, we are unable to reach exactly 4$\times$ less storage. Employing INT8 also runs about 3.5$\times$ faster on average. Note that we estimate this speed-up by collecting the time consumption of each operation and their corresponding speed-up (6$\times$) in INT8, as modern CPUs have limited supports of INT8 arithmetics, e.g., MatMul only. We find that this speed-up is more obvious if the output sequence is longer, e.g., translations in Ro$\rightarrow$En is longer than those in En$\rightarrow$Fr and thus higher speed-up in Ro$\rightarrow$En is observed. This phenomenon arises from the fact that operations that benefit from INT8 such as MatMul occupy a higher portion when generating long sequences, while other fixed time operations such as data preparation become marginal.

\subsection{Analysis}
\label{sec:analysis}

    \begin{table}[t!]
        \centering
        \renewcommand\tabcolsep{4pt}
        \begin{tabular}{l|c|c|c|c|c}
            \hline
            \multicolumn{1}{c|}{\multirow{2}*{System}} & \multicolumn{3}{c|}{BLEU} & \multicolumn{2}{c}{PPL} \\
            \cline{2-4}
            \cline{5-6}
            & En$\rightarrow$Ro & En$\rightarrow$De & En$\rightarrow$Fr & valid & test \\
            \hline
            Baseline & 32.55 & 26.95 & \textbf{40.99} & 29.58 & 31.28 \\
            +Poly & \textbf{32.56} & \textbf{27.13} & 40.90 & \textbf{29.54} & 31.20 \\
            +L1LN & 32.55 & 26.94 & 40.67 & 29.61 & \textbf{31.18} \\
            \hline
        \end{tabular}
        \caption{The ablation study of Integer Transformer.}
        \label{tab:arch-ablation}
    \end{table}

    \begin{table}[t!]
        \centering
        \renewcommand\tabcolsep{3pt}
        \begin{tabular}{l|c|c|c|c|c}
            \hline
            \multicolumn{1}{c|}{\multirow{2}*{System}} & \multicolumn{3}{c|}{BLEU} & \multicolumn{2}{c}{PPL} \\
            \cline{2-4}
            \cline{5-6}
            & En$\rightarrow$Ro & En$\rightarrow$De & En$\rightarrow$Fr & valid & test \\
            \hline
            Ours (FP32) & 32.60 & 27.08 & 40.99 & 29.49 & 30.79 \\
            +$B$ Scale & \textbf{32.54} & 21.84 & 39.45 & 30.88 & 32.25 \\
            +$B \times T$ Scale & \textbf{32.54} & \textbf{26.91} & \textbf{40.00}& \textbf{30.28} & \textbf{31.61} \\
            \hline
        \end{tabular}
        \caption{INT8 performance vs. different sized scales.}
        \label{tab:sp-ablation}
    \end{table}

    We show an ablation study of integer Transformer in Table \ref{tab:arch-ablation}. We can see that replacing the standard attention by polynomial attention generally improve the FP32 result and L1 layer normalization has the close performance to standard layer normalization. These observations imply that either the polynomial attention or the L1 layer normalization is a good alternative to its counterpart in the baseline transformer model.

    Section \ref{sec:bounding} has described how to obtain the initial scale by taking the maximum of the hidden dimension in the FP32 input. This method can be extended to the case of multiple dimensions. In Table \ref{tab:sp-ablation}, we test it on maximizing on $T \times C$ and $C$ given the input of the size $T \times B \times C$, resulting a sized $B$ and $B \times T$ scale respectively. Here $T$ is the input sequence length, $B$ is the batch size and $C$ is the number of the hidden units. The results reveal that using a scale with more entries better preserves the performance, yet the one with fewer entries lowers the computation budget.

\begin{figure}[t!]
        \centering
        \begin{tabular*}{\linewidth}{@{\hskip0pt}r@{\extracolsep{\fill}}r@{\hskip0pt}}
        \multicolumn{2}{@{\hskip0pt}c@{\hskip0pt}}{\hspace*{3pt}\tikz {
            \scriptsize

            \legendmargin=0.16cm
            \legendwidth=0.4cm
            \legendsep=0.1cm

            \coordinate (start) at (0,0);
            \draw[orange,thick,postaction={decorate},decoration={markings,mark=at position 0.5 with {\pgfuseplotmark{*}}}] ([xshift=\legendmargin]start.east) -- +(\legendwidth,0) node[black,right] (l1) {En-Ro};
            \draw[lyyred,thick,postaction={decorate},decoration={markings,mark=at position 0.5 with {\pgfuseplotmark{square*}}}] ([xshift=\legendsep]l1.east) -- +(\legendwidth,0) node[black,right] (l2) {En-De};
            \draw[lyygreen,thick,postaction={decorate},decoration={markings,mark=at position 0.5 with {\pgfuseplotmark{diamond*}}}] ([xshift=\legendsep]l2.east) -- +(\legendwidth,0) node[black,right] (l3) {En-Fr};
            \draw[lyyblue,thick,postaction={decorate},decoration={markings,mark=at position 0.5 with {\pgfuseplotmark{triangle*}}}] ([xshift=\legendsep]l3.east) -- +(\legendwidth,0) node[black,right] (l4) {WikiText-103};
            \coordinate (end) at ([xshift=\legendmargin-3pt]l4.east);
            \begin{pgfonlayer}{background}
            \node[rectangle,draw,inner sep=1pt] [fit = (start) (l1) (l2) (l3) (l4) (end)] {};
            \end{pgfonlayer}
        }}\\[0pt]
        \begin{tikzpicture}
          \begin{axis}[
            width=0.45\linewidth,height=0.5\linewidth,
            yticklabel style={/pgf/number format/fixed,/pgf/number format/precision=1},
            ymin=-0.7,ymax=0.4,
            enlarge y limits=0.1,
            ytick={-0.7,-0.425,...,0.4},
            ylabel={$\triangle$BLEU},
            ylabel near ticks,
            xlabel={$n$},
            xlabel near ticks,
            symbolic x coords={1,2,3,4,5},
            xmajorgrids=true,
            ymajorgrids=true,
            grid style=dashed,
            xtick=data,
            every tick label/.append style={font=\tiny},
            xlabel style={yshift=-0.05cm},
            label style={font=\scriptsize},
          ]
            \addplot [orange,thick,mark=*] coordinates {
                (1,-0.62) (2,-0.16) (3,0.05) (4,0.1) (5,0.2)
            };
            \addplot [lyyred,thick,mark=square*] coordinates {
                (1,0.06) (2,0.14) (3,0.29) (4,0.12) (5,-0.05)
            };
            \addplot [lyygreen,thick,mark=diamond*] coordinates {
                (1,-0.29) (2,-0.06) (3,0.03) (4,-0.19) (5,0.01)
            };
          \end{axis}
          \begin{axis}[
            width=0.45\linewidth,height=0.5\linewidth,
            yticklabel style={/pgf/number format/fixed,/pgf/number format/precision=1},
            ymin=-0.4,ymax=0.6,
            enlarge y limits=0.1,
            xtick=\empty,
            ytick={-0.4,-0.15,...,0.6},
            label style={font=\scriptsize},
            axis y line*=right,
            xlabel near ticks,
            symbolic x coords={1,2,3,4,5},
            every tick label/.append style={font=\tiny},
            xlabel style={yshift=-0.05cm},
          ]
            \addplot [lyyblue,thick,mark=triangle*] coordinates {
                (1,-0.33) (2,0.10) (3,0.49) (4,0.18) (5,0.33)
            };
          \end{axis}
        \end{tikzpicture}
        &
        \begin{tikzpicture}
          \begin{axis}[
            width=0.45\linewidth,height=0.5\linewidth,
            yticklabel style={/pgf/number format/fixed,/pgf/number format/precision=0},
            ymin=-12.0,ymax=0,
            enlarge y limits=0.1,
            ytick={-12.0,-9.6,...,0},
            xlabel={\#Bits},
            xlabel near ticks,
            symbolic x coords={6,7,8,9,10},
            xmajorgrids=true,
            ymajorgrids=true,
            grid style=dashed,
            xtick=data,
            every tick label/.append style={font=\tiny},
            label style={font=\scriptsize},
          ]
            \addplot [orange,thick,mark=*] coordinates {
                (6,-0.24) (7,-0.04) (8,-0.01) (9,0.09) (10,0.04)
            };
            \addplot [lyyred,thick,mark=square*] coordinates {
                (6,-3.64) (7,-0.51) (8,-0.07) (9,0.08) (10,0.05)
            };
            \addplot [lyygreen,thick,mark=diamond*] coordinates {
                (6,-11.46) (7,-3.10) (8,-0.88) (9,-0.40) (10,-0.32)
            };
          \end{axis}
          \begin{axis}[
            width=0.45\linewidth,height=0.5\linewidth,
            yticklabel style={/pgf/number format/fixed,/pgf/number format/precision=1},
            ymin=-50.0,ymax=0.0,
            enlarge y limits=0.1,
            xtick=\empty,
            ytick={-50.0,-40.0,...,0.0},
            ylabel={$\triangle$PPL},
            ylabel near ticks,
            axis y line*=right,
            symbolic x coords={6,7,8,9,10},
            every tick label/.append style={font=\tiny},
            label style={font=\scriptsize},
          ]
            \addplot [lyyblue,thick,mark=triangle*] coordinates {
                (6,-38.65) (7,-3.68) (8,-0.43) (9,0.21) (10,0.34)
            };
          \end{axis}
        \end{tikzpicture}
        \\[0pt]
        \end{tabular*}
        \caption{Sensitivity analysis ($n$: the degree of the polynomial function; \#Bits: the number of bits used in the inference).}
        \label{fig:sensitive}
    \end{figure}
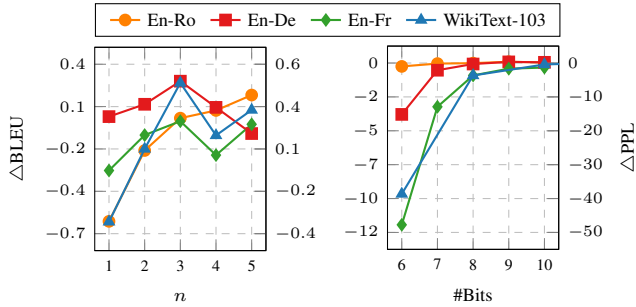

    Also, we plot how hyper-parameters relate to performance. We can see from the left of Fig. \ref{fig:sensitive} that $n>1$ results in much better performance than $n=1$ in all tasks, indicating the necessity of non-linearity in the attention. But higher $n$ does not necessarily lead to better results, where $n=3$ performs the best in En$\rightarrow$De and WikiText-103. The right of Fig. \ref{fig:sensitive} shows that adding a few bits can recover most of the performance, especially for those suffer from great loss in INT8 inference, e.g., En$\rightarrow$Fr. Moreover, we observe that the performance of En$\rightarrow$Ro decreases slightly with 6 bit, which suggests that further speed-up might be available.

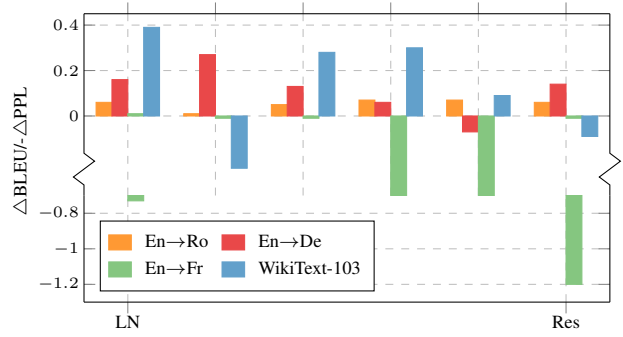
\begin{figure}[t!]
        \centering
        \begin{tabular}{@{\hskip0pt}r@{\hskip3pt}l@{\hskip0pt}}
            {\rotatebox{90}{\hspace*{1.5cm}\scriptsize$\triangle$BLEU/-$\triangle$PPL}}
            &
            \begin{tikzpicture}
                \begin{groupplot}[
                    width=\linewidth,
                    yticklabel style={/pgf/number format/fixed,/pgf/number format/precision=2},
                    symbolic x coords={Emb,Attn,FFN,LN,Res,Proj},
                    enlarge x limits=0.1,
                    ylabel near ticks,
                    ybar=0pt,
                    xtick=data,
                    ytick=data,
                    xticklabel style={font=\scriptsize,inner sep=0pt,outer sep=1pt},
                    yticklabel style={font=\tiny,inner sep=0pt,outer sep=2pt},
                    xmajorgrids=true,
                    ymajorgrids=true,
                    grid style=dashed,
                    legend image code/.code={
                        \draw [#1] (0cm,-0.1cm) rectangle (0.3cm,0.1cm);
                    },
                    legend style={font=\scriptsize,at={(0.03,0.10)},anchor=south west,inner sep=3pt,column sep=3pt},
                    legend cell align={left},
                    legend columns=2,
                    legend image post style={anchor=east},
                    group style={
                        group size=1 by 2,
                        xticklabels at=edge bottom,
                        vertical sep=0pt,
                    },
                ]
                    \nextgroupplot[
                        ymin=-0.35,ymax=0.45,
                        height=4cm,
                        bar width=6pt,
                        axis y discontinuity=crunch,
                        axis x line*=top,
                        ytick={0,0.2,...,0.4},
                    ]
                        \addplot [draw=orange!80!white,fill=orange!80!white] coordinates {(Emb,0.06) (Attn,0.01) (FFN,0.05) (LN,0.07) (Res,0.07) (Proj,0.06)};\label{enro}
                        \addplot [draw=lyyred!80!white,fill=lyyred!80!white] coordinates {(Emb,0.16) (Attn,0.27) (FFN,0.13) (LN,0.06) (Res,-0.07) (Proj,0.14)};\label{ende}
                        \addplot [draw=lyygreen!60!white,fill=lyygreen!60!white] coordinates {(Emb,0.01) (Attn,-0.01) (FFN,-0.01) (LN,-0.5) (Res,-0.5) (Proj,-0.01)};
                        \addplot [draw=lyyblue!70!white,fill=lyyblue!70!white] coordinates {(Emb,0.39) (Attn,-0.23) (FFN,0.28) (LN,0.30) (Res,0.09) (Proj,-0.09)};\label{wiki}
                    \nextgroupplot[
                        ymin=-1.3,ymax=-0.7,
                        height=3cm,
                        bar width=6pt,
                        axis x line*=bottom,
                        ytick={-1.2,-1,...,-0.8},
                    ]
                        \addlegendimage{/pgfplots/refstyle=enro,xshift=3pt}\addlegendentry{En$\rightarrow$Ro}
                        \addlegendimage{/pgfplots/refstyle=ende}\addlegendentry{En$\rightarrow$De}
                        \addplot [draw=lyygreen!60!white,fill=lyygreen!60!white,xshift=0.5*\pgfplotbarwidth] coordinates {(Emb,0.01) (Attn,-0.01) (FFN,-0.01) (LN,-0.73) (Res,-1.20) (Proj,-0.01)};\addlegendentry{En$\rightarrow$Fr}
                        \addlegendimage{/pgfplots/refstyle=wiki}\addlegendentry{WikiText-103}
                \end{groupplot}
            \end{tikzpicture}\\[0pt]
        \end{tabular}
        \caption{Performance improvement ($>0$) and loss ($<0$) of applying INT8 to modules (Emb: the embedding; Attn: the attention; FFN: the feed-forward network; LN: the layer normalization; Res: the residual connection; Proj: the output projection.)}
        \label{fig:sp-ablation}
    \end{figure}

    We next investigate which factor has a significant impact on the performance by presenting details on which module is responsible for the INT8 performance loss. As can be seen in Fig. \ref{fig:sp-ablation}, if the performance drop is not significant, each module contribute similarly, otherwise a few modules should be blamed for. This fact suggests that poor INT8 performance is mainly led by one or two crucial points, e.g., the layer normalization and the residual connection in En$\rightarrow$Fr.

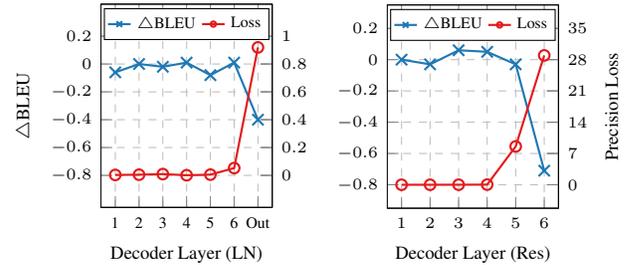
\begin{figure}[t!]
        \centering
        \hspace*{\fill}
        \begin{tikzpicture}
            \begin{axis}[
                width=0.45\linewidth,height=0.5\linewidth,
                yticklabel style={/pgf/number format/fixed,/pgf/number format/precision=1},
                enlarge x limits=0.1,
                enlarge y limits=0.23,
                ymin=-0.8,ymax=0.2,
                ylabel={$\triangle$BLEU},
                ylabel near ticks,
                xlabel={Decoder Layer (LN)},
                xlabel near ticks,
                symbolic x coords={1,2,3,4,5,6,Out},
                xmajorgrids=true,
                ymajorgrids=true,
                grid style=dashed,
                xtick=data,
                ytick={-0.8,-0.6,...,0.2},
                every tick label/.append style={font=\tiny},
                label style={font=\scriptsize,align=center},
            ]
                \addplot [lyyblue,thick,mark=x,mark options={scale=1.5,solid}] coordinates {
                    (1,-0.06) (2,0.00) (3,-0.02) (4,0.01) (5,-0.08) (6,0.01) (Out,-0.4)
                };\label{ln}
            \end{axis}
            \begin{axis}[
              width=0.45\linewidth,height=0.5\linewidth,
              yticklabel style={/pgf/number format/fixed,/pgf/number format/precision=1},
              ymin=0.0,ymax=1,
              enlarge x limits=0.1,
              enlarge y limits=0.23,
              xtick=\empty,
              ytick={0,0.2,...,1},
              label style={font=\scriptsize},
              axis y line*=right,
              xlabel near ticks,
              symbolic x coords={1,2,3,4,5,6,Out},
              every tick label/.append style={font=\tiny},
              xlabel style={yshift=-0.05cm},
              legend style={font=\tiny,at={(0.5,0.98)},anchor=north,inner sep=1pt},
              legend image post style={scale=0.5},
              legend columns=2,
              legend cell align={left},
            ]
                \addlegendimage{/pgfplots/refstyle=ln}\addlegendentry{$\triangle$BLEU}
                \addplot [lyyred,thick,mark=o,mark options={scale=1,solid}] coordinates {
                    (1,3.22e-03) (2,4.88e-03) (3,9.28e-03) (4,1.01e-03) (5,5.33e-03) (6,0.0524835) (Out,0.9191021)
                };
                \addlegendentry{Loss}
            \end{axis}
        \end{tikzpicture}
        \hfill
        \begin{tikzpicture}
            \begin{axis}[
                width=0.45\linewidth,height=0.5\linewidth,
                yticklabel style={/pgf/number format/fixed,/pgf/number format/precision=1},
                enlarge x limits=0.1,
                enlarge y limits=0.15,
                ymin=-0.8,ymax=0.2,
                xlabel={Decoder Layer (Res)},
                xlabel near ticks,
                xmajorgrids=true,
                ymajorgrids=true,
                grid style=dashed,
                xtick=data,
                ytick={-0.8,-0.6,...,0.2},
                every tick label/.append style={font=\tiny},
                label style={font=\scriptsize,align=center},
            ]
                \addplot [lyyblue,thick,mark=x,mark options={scale=1.5,solid}] coordinates {
                    (1,-0.00) (2,-0.03) (3,0.06) (4,0.05) (5,-0.03) (6,-0.71)
                };\label{res}
            \end{axis}
            \begin{axis}[
              width=0.45\linewidth,height=0.5\linewidth,
              yticklabel style={/pgf/number format/fixed,/pgf/number format/precision=1},
              ymin=0.0,ymax=35.0,
              enlarge x limits=0.1,
              enlarge y limits=0.15,
              xtick=\empty,
              ytick={0,7,...,35},
              ylabel={Precision Loss},
              ylabel near ticks,
              label style={font=\scriptsize},
              axis y line*=right,
              xlabel near ticks,
              symbolic x coords={1,2,3,4,5,6},
              every tick label/.append style={font=\tiny},
              xlabel style={yshift=-0.05cm},
              legend style={font=\tiny,at={(0.5,0.98)},anchor=north,inner sep=1pt},
              legend image post style={scale=0.5},
              legend columns=2,
              legend cell align={left},
            ]
                \addlegendimage{/pgfplots/refstyle=res}\addlegendentry{$\triangle$BLEU}
                \addplot [lyyred,thick,mark=o,mark options={scale=1,solid}] coordinates {
                    (1,0.001802167) (2,0.005030057) (3,0.012637655) (4,0.034545465) (5,8.579899222) (6,28.95547939)
                };
                \addlegendentry{Loss}
            \end{axis}
        \end{tikzpicture}
        \hspace*{\fill}
        \caption{Precision loss vs. Performance loss (En$\rightarrow$Fr, Out: the last layer normalization before the output projection).}
        \label{fig:loss}
      \end{figure}

    As implied by Fig. \ref{fig:sp-ablation}, we make an in-depth analysis to see whether the high precision loss connects to the poor performance of applying INT8 to the layer normalization and the residual connection in En$\rightarrow$Fr. To evaluate the precision loss, we choose the mean square error between the FP32 activations and the de-quantized INT8 ones as the proxy. Fig. \ref{fig:loss} shows that there exists a positive relationship between precision loss and performance loss, i.e., $\triangle$BLEU. Interestingly, most loss occurs in the last layer. Noting that the residual connection is the sum of all outputs of the residual branches in previous layers, the last residual connection will produce the result with large values, which might suffer from greater precision loss through the quantization.

\section{Related Work}

    Employing the low precision data type for neural networks to accelerate the network computation or save storage space has a long history. Early work has shown that training and inference with the ternary (2-bit) or even binary (1-bit) network is possible \cite{DBLP:conf/nips/HubaraCSEB16}. But these results have restricted to simple architectures, such as the feed-forward networks. Recent work mainly focuses on training a sophisticated network with higher precision, such as 32-bit (FP32) and 16-bit floating point (FP16) \cite{DBLP:conf/iclr/MicikeviciusNAD18} but attempts to inference with fewer bits, such as 8-bit fixed point (INT8) \cite{DBLP:conf/cvpr/JacobKCZTHAK18}. However, most of them have limited to computer vision and only a few of them discuss how to leverage low precision to infer the complicate Transformer model in natural language processing.

    \citet{DBLP:journals/corr/abs-1906-00532} first demonstrates that Transformer can be inferred with INT8. But some operations are still performed in FP32 and its INT8 performance is not evaluated by common metrics, e.g., BLEU. Though more recent work \cite{DBLP:journals/corr/abs-1910-10485,DBLP:journals/corr/abs-2001-00926} share the same limitation of partially relying on FP32, they report better INT8 results by tailoring the training as well as the quantization method to the Transformer model. This work, on the other hand, takes a step toward fully INT8 inference without any FP32 operation for the Transformer model. The forward propagation flows purely on INT8 and shows competitive performance without modifying the training process.

\section{Conclusion}

    In this work, we present an (almost) fully INT8 inference algorithm \emph{Scale Propagation}, which propagates the INT8 tensor and its scale to resolve the scale incompatibility problem. Moreover, we propose \emph{Integer Transformer} to address the INT8 incompatibility issue in the Transformer model, which replaces the exponential function and the square root function by the polynomial function and the absolute value function respectively. Our experiments show that our method achieves competitive INT8 performance in machine translation and language modelling tasks.

\section*{Acknowledgments}

    This work was supported in part by the National Science Foundation of China (Nos. 61876035 and 61732005), the National Key R\&D Program of China (No. 2019QY1801) and the Opening Project of Beijing Key Laboratory of Internet Culture and Digital Dissemination Research. The authors would like to thank anonymous reviewers for their comments.


\bibliographystyle{named}
\bibliography{ijcai20}

\end{document}